%% file: scs23paper.tex
\documentclass{scspaperproc}

\usepackage{latexsym}
\usepackage{graphicx}
\usepackage{mathptmx}
\usepackage{subcaption}
\usepackage{amsmath}
\usepackage{amsfonts}
\usepackage{amssymb}
\usepackage{amsbsy}
\usepackage{amsthm}
\usepackage{hyphenat}
\newtheoremstyle{scsthe}
{8pt}
{8pt}
{\it}
{}
{\bf}
{.}
{.5em}
{}

\theoremstyle{scsthe}

\begin{document}

%
%


\pagestyle{fancyplain}

\thispagestyle{plain}
\firstPageHead{}

\chead{\fancyplain{}{\itshape\small Deshkar, Kshirsagar, Hayatnagarkar, and Venugopalan \vspace{8pt}}}

\rhead{}
\cfoot{}
\renewcommand{\headrulewidth}{0pt} 

\input{scsprocbib.tex}           

\setlength{\baselineskip}{12.7pt}

\def\SCSconferenceacro{ANNSIM'23}

\def\SCSpublicationyear{2023}

\def\SCSconferencedates{May 23-26}

\def\SCSconferencevenue{Mohawk College, ON, CANADA}

\title{Epidemic Control on a Large-Scale-Agent-Based Epidemiology Model using Deep Deterministic Policy Gradient}

\author{
\\
Gaurav Deshkar\\
Jayanta Kshirsagar\\
Harshal Hayatnagarkar\\
Janani Venugopalan\\ [12pt]
Engineering for Research (e4r) Thoughtworks Technologies, Pune, India\\
\{gauravd, jayantak, harshalh, janani.venugopalan\}@thoughtworks.com
}


\maketitle

\section*{Abstract}

To mitigate the impact of the pandemic, several measures include lockdowns, rapid vaccination programs, school closures, and economic stimulus. These interventions can have positive or unintended negative consequences. Current research to model and determine an optimal intervention automatically through round-tripping is limited by the simulation objectives, scale (a few thousand individuals), model types that are not suited for intervention studies, and the number of intervention strategies they can explore (discrete vs continuous). We address these challenges using a Deep Deterministic Policy Gradient (DDPG) based policy optimization framework on a large-scale (100,000 individual) epidemiological agent-based simulation where we perform multi-objective optimization. We determine the optimal policy for lockdown and vaccination in a minimalist age-stratified multi-vaccine scenario with a basic simulation for economic activity. With no lockdown and vaccination (mid-age and elderly), results show optimal economy (individuals below the poverty line) with balanced health objectives (infection, and hospitalization). An in-depth simulation is needed to further validate our results and open-source our framework.

\textbf{Keywords:} Policy Informatics, Large Scale Simulation, Optimization, DDPG, Agent-Based Simulation.

\section{Introduction}
\input{sections/introduction.tex}

\section{Key Contributions}
\input{sections/key_contribution.tex}
\section{Methods}
\input{sections/methods.tex}

\section{Results}
\input{sections/results.tex}

\section{Discussion}
\input{sections/discussion.tex}
\section{Conclusion}
\input{sections/conclusion.tex}

\bibliographystyle{scsproc}
\bibliography{mybibfile}

\end{document}

%% file: scsprocbib.tex
\makeatletter
\let\@internalcite\cite
\def\cite{\def\@citeseppen{-1000}%
    \def\@cite##1##2{(##1\if@tempswa , ##2\fi)}%
    \def\citeauthoryear##1##2##3{##1 ##3}\@internalcite}
\def\citeNP{\def\@citeseppen{-1000}%
    \def\@cite##1##2{##1\if@tempswa , ##2\fi}%
    \def\citeauthoryear##1##2##3{##1 ##3}\@internalcite}
\def\citeN{\def\@citeseppen{-1000}%
    \def\@cite##1##2{##1\if@tempswa, ##2)\else{}\fi}%
    \def\citeauthoryear##1##2##3{##1 (##3)}\@citedata}
\def\citeA{\def\@citeseppen{-1000}%
    \def\@cite##1##2{(##1\if@tempswa , ##2\fi)}%
    \def\citeauthoryear##1##2##3{##1}\@internalcite}
\def\citeANP{\def\@citeseppen{-1000}%
    \def\@cite##1##2{##1\if@tempswa , ##2\fi}%
    \def\citeauthoryear##1##2##3{##1}\@internalcite}
\def\shortcite{\def\@citeseppen{-1000}%
    \def\@cite##1##2{(##1\if@tempswa , ##2\fi)}%
    \def\citeauthoryear##1##2##3{##2 ##3}\@internalcite}
\def\shortciteNP{\def\@citeseppen{-1000}%
    \def\@cite##1##2{##1\if@tempswa , ##2\fi}%
    \def\citeauthoryear##1##2##3{##2 ##3}\@internalcite}
\def\shortciteN{\def\@citeseppen{-1000}%
    \def\@cite##1##2{##1\if@tempswa, ##2\else{}\fi}%
    \def\citeauthoryear##1##2##3{##2 (##3)}\@citedata}
\def\shortciteA{\def\@citeseppen{-1000}%
    \def\@cite##1##2{(##1\if@tempswa , ##2\fi)}%
    \def\citeauthoryear##1##2##3{##2}\@internalcite}
\def\shortciteANP{\def\@citeseppen{-1000}%
    \def\@cite##1##2{##1\if@tempswa , ##2\fi}%
    \def\citeauthoryear##1##2##3{##2}\@internalcite}
\def\citeyear{\def\@citeseppen{-1000}%
    \def\@cite##1##2{(##1\if@tempswa , ##2\fi)}%
    \def\citeauthoryear##1##2##3{##3}\@citedata}
\def\citeyearNP{\def\@citeseppen{-1000}%
    \def\@cite##1##2{##1\if@tempswa , ##2\fi}%
    \def\citeauthoryear##1##2##3{##3}\@citedata}
%
%
%
\def\@citedata{%
    \@ifnextchar [{\@tempswatrue\@citedatax}%
                  {\@tempswafalse\@citedatax[]}%
}

\def\@citedatax[#1]#2{%
\if@filesw\immediate\write\@auxout{\string\citation{#2}}\fi%
  \def\@citea{}\@cite{\@for\@citeb:=#2\do%
    {\@citea\def\@citea{, }\@ifundefined
       {b@\@citeb}{{\bf ?}%
       \@warning{Citation `\@citeb' on page \thepage \space undefined}}%
{\csname b@\@citeb\endcsname}}}{#1}}%

%
\def\@citex[#1]#2{%
\if@filesw\immediate\write\@auxout{\string\citation{#2}}\fi%
  \def\@citea{}\@cite{\@for\@citeb:=#2\do%
    {\@citea\def\@citea{, }\@ifundefined
       {b@\@citeb}{{\bf ?}%
       \@warning{Citation `\@citeb' on page \thepage \space undefined}}%
{\csname b@\@citeb\endcsname}}}{#1}}%

%
\def\@biblabel#1{}
\makeatother

\newdimen\bibindent
\bibindent=.25in

\def\thebibliography#1{\section*{\refname}\list
   {}{\settowidth\labelwidth{[#1]}
   \leftmargin \bibindent
   \itemindent -\bibindent
   \listparindent \itemindent
	 \itemsep 4pt
   \parsep 0pt
   \usecounter{enumi}}
   \def\newblock{}
   \sloppy
   \sfcode`\.=1000\relax}

%% file: sections/introduction.tex
\label{sec:intro}
The COVID-19 pandemic has caused more than 460 million infections, 6 million deaths (till date), and a 3\% decrease in world GDP in 2020 alone ~\shortcite{wu2022economic}. To mitigate the impact of the pandemic, policymakers world over have come up with several measures including lockdowns, rapid vaccination programs, school closures, public-transport closures, and economic stimulus ~\shortcite{perra2021non}. Some of these interventions, such as prolonged lockdowns and public transport closures have been shown to have severe consequences ~\shortcite{bavli2020harms} including an increase in economic disparities and poor health outcomes due to overcrowding. It is crucial to have efficient and reliable optimization methods to guide our decision-making.  

Until very recently, despite a significant body of literature on epidemiological models (survey papers ~\shortcite{shankar2021systematic} and books ~\shortcite{martcheva2015introduction}), policy informatics to control the effects of such epidemics and to mitigate the effects of policy interventions is relatively nascent (~\shortciteNP{puron2016opportunities}, ~\shortciteNP{veenstra2017data}, ~\shortciteNP{wan2022advances}). The recent work on the automatic control of epidemics and optimization has focused on the use of either greedy algorithms ~\shortcite{minutoli2020preempt} or reinforcement learning frameworks such as Deep Q-learning variants (~\shortciteNP{khadilkar2020optimising}, ~\shortciteNP{ohi2020exploring}, ~\shortciteNP{colas2021epidemioptim}, ~\shortciteNP{bampa2022learning}). Techniques such as Q-learning and greedy optimization can only choose from a predetermined set of intervention strategies (states and actions) e.g. high vs low vaccination; choose one among $n$ types of masks. In a realistic public health scenario, these techniques cannot be used to model and optimize continuous factors such as the number of vaccinations needed and the duration of lockdowns ~\shortcite{lillicrap2015continuous}. Studies using continuous reinforcement learning frameworks such as Deep Deterministic Policy Gradient (DDPG), Twin Delayed DDPG, and Proximal Policy Optimization (PPO) ~\shortcite{chadi2022reinforcement} have been proposed to handle such continuous actions.  However, these studies are limited in the simulation objectives (only one or two of vaccination, lockdown, economy), scale (few thousand individuals), or use equation-based models which are not suited for intervention studies. This is very important since the use of inappropriate models or optimization algorithms can have unintended and often negative consequences ~\shortcite{lunich2021using}.

We overcome these challenges and demonstrate a Deep Deterministic Policy Gradient (DDPG) ~\shortcite{hou2017novel} based policy optimization framework on a large-scale epidemiological simulation where we optimize for multiple objectives (health and economy). We determine the optimal policy for lockdown and vaccination in an age-stratified multi-vaccine scenario with a basic simulation for economic activity. To facilitate the further extension of our study in terms of scale, multiple objectives, and improved simulation models, we plan to open-source our code, the simulation, the optimization methodology, and the hyper-parameters used in our study.

%% file: sections/key_contribution.tex
The key contributions of this study are as follows:
\begin{itemize}
    \item We have developed an agent-based epidemiological model for a population of 100,000 individuals (agents) where we simulate a lockdown, age-stratified multi-vaccine scenario, and a basic economic model (while our model can simulate multiple scenarios, in this study we have chosen these two to illustrate our results). Using this model, we demonstrate the use of a reinforcement learning algorithm ``DDPG" to automatically discover the optimal policy to balance the economy and human lives. This allows for the optimization of continuous factors as opposed to purely discrete ones. There is also a significant increase in the scale of the model over existing policy optimization literature ~\shortcite{chadi2022reinforcement}. Since epidemics are heavily influenced by population density, we have started with a city (due to constraints on computation). In the near future, we will expand to include large cities and a combination of cities and rural areas. 
    \item The policy can account for user preferences on the optimization of health and economic costs, with a weight on each factor.
    \item The algorithm can account for any changes in disease parameters, and cost definitions. Only the relevant input values need to be updated by the user.
\end{itemize}

%% file: sections/methods.tex
In this study, we perform a policy optimization analysis of an epidemic control model using reinforcement learning (RL). 
\subsection{Epidemiological Model}
The model presented in the paper is a variation of the SEIR model ~\shortcite{chitnis2017introduction}. In this agent-based model, we model 100,000 agents which is a factor-fold increase from similar optimization literature ~\shortcite{chadi2022reinforcement}. We chose this number since 99\% of the US (sparsely populated country) cities ~\shortcite{duffin2019number} and a significant portion of Indian cities (densely populated country) ~\shortcite{pradhan2017unacknowledged} is smaller than 100,000 people each. With a good representation in terms of population, our model consists of four key aspects
\begin{itemize}
    \item \textbf{Individuals}: In the model, we have two different types of individuals (agents). Agents with an age greater than the age of 30 are considered employed and all agents below the age of 30 are considered to be students. Every agent follows a schedule, where they spend 12 hours(1 vector tick) at home followed by 12 hours at either the office or school based on their age. We chose these locations as to demonstrate a minimalist model where people are moving.
    \item \textbf{Geography}: Geography consists of four main areas, which are Houses, Offices, Schools, and Hospitals. Based on the schedule and its attributes, an agent would be at home or office/school at any given point in time. Hospitalized individuals will spend all their time in the hospital.
    \item \textbf{Disease dynamics}: For modeling the infectious disease we are using a 9-compartment disease model which is the SEIR model ~\shortcite{hazra2021indsci}. Every agent would be at exactly one and only one stage at any point in time. The disease compartments are shown in ``Fig.~\ref{fig_disease_flow}'' ~\shortcite{hazra2021indsci}. The transition factors ($\beta , 1-\gamma , 1-\delta, \sigma$) are age-stratified as shown in ``Table.~\ref{tab_disease_age_stratified}'' (~\shortciteNP{kerr2021covasim}, ~\shortciteNP{hazra2021indsci}). The duration in terms of the number of days spent in a particular compartment is derived from a log-normal distribution with the means and standard deviations shown in ``Table.~\ref{tab_duration_by_compartment}'' (~\shortciteNP{kerr2021covasim}, ~\shortciteNP{hazra2021indsci}, ~\shortciteNP{childs2020impact}).
    \begin{figure}[htbp]
    \centerline{\includegraphics[scale=0.3]{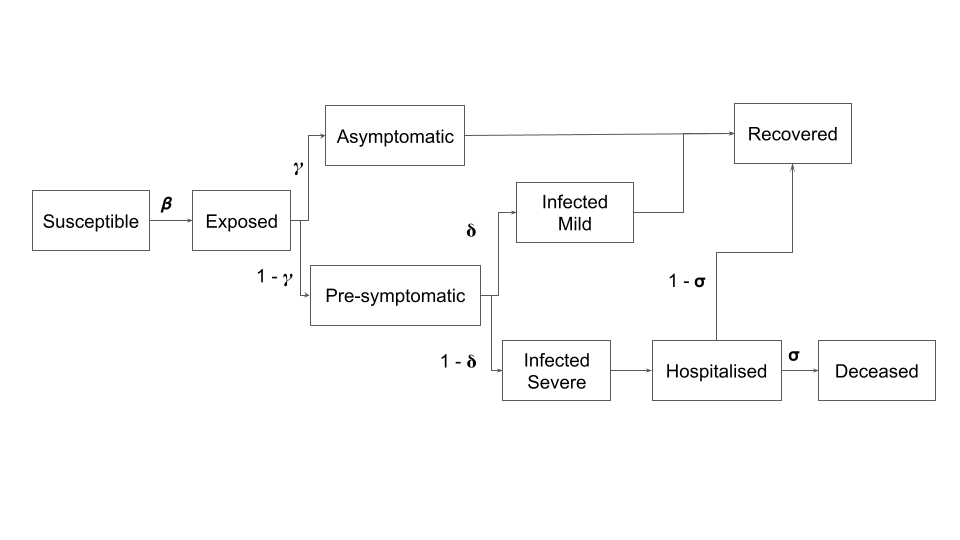}}
    \caption{Disease compartment model.}
    \label{fig_disease_flow}
    \end{figure}

    \begin{table}[htbp]
        \caption{Disease transition factor by age group}
        \begin{center}
        \begin{tabular}{|c|c|c|c|c|}
        \hline
        \textbf{Age Group} & \textbf{\textit{ $\beta$ multiplier}}& \textbf{\textit{ 1 - $\gamma$ }}  & \textbf{\textit{ 1 - $\delta$ }}  & \textbf{\textit{ $\sigma$ }} \\
        \hline
            0 - 9 & 0.34 & 0.5  & 0.0005  &  0.00002  \\
        \hline
            10 - 19 & 0.67 & 0.55  & 0.00165  & 0.00002  \\
        \hline 
            20 - 29 & 1.0 & 0.6   & 0.00720 & 0.0001  \\
        \hline 
            30 - 39 & 1.0 & 0.65   & 0.02080 & 0.00032  \\    
        \hline 
            40 - 49 & 1.0 & 0.7   & 0.03430 &  0.00098  \\  
        \hline 
            50 - 59 & 1.0 & 0.75   & 0.07650 & 0.00265  \\ 
        \hline 
            60 - 69 & 1.0 & 0.8   & 0.13280 & 0.00766  \\  
        \hline 
            70 - 79 & 1.24 & 0.85   & 0.20655 & 0.02439  \\   
        \hline 
            80 - 89 & 1.47 & 0.9   & 0.24570 & 0.08292 \\  
        \hline 
            90 - 99 & 1.47 & 0.9   & 0.24570 & 0.16190  \\      
        \hline 
       
        \multicolumn{5}{l}{ \textbf{Note:} $\beta$ defined as 0.5 is multiplied by "$\beta$ multiplier" for age.}
        \end{tabular}
        \label{tab_disease_age_stratified}
        \end{center}
    \end{table}
    
    \begin{table}[htbp]
        \caption{Number of days spent in compartment}
        \begin{center}
        \begin{tabular}{|c|c|c|}
        \hline
        \textbf{Compartment} & \textbf{\textit{ Mean}}& \textbf{\textit{ Standard Deviation  }} \\
        \hline
            Exposed & 4.5  & 1.5  \\
        \hline
            Asymptomatic & 8.0  & 2.0  \\    
        \hline
            Pre-Symptomatic & 1.1 & 0.9  \\
        \hline
            Infected Mild & 8.0  & 2.0  \\
        \hline
            Infected Severe & 1.5 & 2.0 \\  
        \hline
            Hospitalized & 18.1 & 6.3\\  
        \hline 
        \multicolumn{3}{l}{ \textbf{Note:} derived from log normal distribution.}
        \end{tabular}
        \label{tab_duration_by_compartment}
        \end{center}
    \end{table}
    
    \item \textbf{Economy}: We model a basic economic activity  with the house as the central unit. Each house has some savings, daily income, and daily expenses. Initial savings are distributed by a normal distribution with a mean of 500 and a standard deviation of 350. The daily income is also distributed by a normal distribution with a mean of 100 and a standard deviation of 30. The daily expense per person is 10 units. A house has about 4 people and one of them is designated as the household head. When the household head is healthy, the daily expenses are deducted from the daily income, and the rest is added to the savings of the house. When the household head is following a lockdown or becomes sick or is deceased, the income stops, and the daily expenses are deducted from the house savings. Agents/people are marked below the poverty line based on the savings of the house they belong to. The poverty line for our model is 100 economic units.
\end{itemize}

To prevent the excessive spread of the epidemic, certain interventions such as lockdowns are imposed. The model, in this study, has the following interventions:
\begin{itemize}
    \item \textbf{Lockdown}: Most individuals stay home when the lockdown is applied. Only essential workers are allowed to work. There are also some lockdown violators. In our model, about 20\% of the population are essential workers and about 10\% are lockdown violators.
    \item \textbf{Vaccination}: When an individual has been vaccinated their chances of getting infected ($\beta$) are reduced based on the vaccine's effectiveness. A vaccinated individual's $\gamma$ is increased by 80\%. i.e. in case of infection, it is more likely for them to get an asymptomatic infection. The individual disease transmission factor is reduced by 20\%, which implies that they are less likely to pass on the disease. There are two types of vaccines in our model, with tunable effectiveness. The population is age-stratified into three age groups 0-17, 18-59, and 60-99. 
\end{itemize}
All the parameters in the model are picked from epidemiological literature. 

\subsection{Optimization of Interventions}
In this study, the factors we optimize are as follows:
\begin{itemize}
    \item \textbf{Lockdown}: $StartDay$ and $EndDay$
    \item \textbf{Vaccination}: $StartDay$ and $EndDay$ for each of the three age stratification.
\end{itemize}

Since each of these values are continuous and can accept any value from the start of day 0 to the end of day 100, we decided to use a DDPG-based reinforcement learning framework for our optimization. As opposed to previous DQN-based techniques, which could handle only a set of discrete optimization states and intervention actions, DDPG can work with continuous optimization states and intervention actions.
Deep Deterministic Policy Gradient (DDPG) ~\shortcite{hou2017novel} is a reinforcement learning technique that combines both Q-learning ~\shortcite{watkins1992q} and Policy gradients ~\shortcite{silver2014deterministic}. DDPG consists of two models: Actor and Critic. The actor takes the state (continuous) as the input and outputs the exact action (continuous), instead of a probability distribution over actions. The critic takes in the state and action as the inputs and outputs of the probable Q-value. DDPG is an “off”-policy method and is used in the continuous action setting. Since the DDPG actor computes the action directly instead of a probability distribution over actions (leading to discrete action spaces), we have used it for our optimization. 

\subsection{Evaluation Criteria}
For each experiment, we evaluate the best and worst-case baseline scenarios with respect to the number of infections, hospitalizations, deaths, and below-poverty-line individuals. The baselines we ran were as follows : 
\begin{enumerate}
    \item No Lockdown, No Vaccination
    \item Full Lockdown, Vaccinations through Day 1-100
    \item No Lockdown, Vaccinations through Day 1-100
    \item Lockdown for 30 days initially, Vaccinations through Day 1-100
\end{enumerate}
We compared the optimization results with each of these test cases.

%% file: sections/results.tex
To test the sanity of the  optimization framework, we first ran simplistic scenarios where lockdowns had no economic impact and the vaccinations were perfect. In such a scenario, the model always imposed a 100\% lockdown and up to 60\% vaccination. Following this, we ran our experimental set-up described below.

\subsection{Experimental Setup}
We tested for a few different experimental setups including high and low initial infection experiments with low and high availability of efficient vaccines (``Table \ref{experiments}"). In all of these experiments, up to 90\% of the population can be vaccinated. As mentioned above, we optimized on lockdowns and the vaccination starts and stop. 

    \begin{table}[htbp]
        \caption{Optimization Experiments Executed}
        \begin{center}
        \begin{tabular}{|c|c|c|c|c|}
        \hline
        \textbf{Parameters} & \textbf{Exp 1}& \textbf{Exp 2}  & \textbf{Exp 3} & \textbf{Exp 4} \\
        \hline
         Initial Infections Percent & 15 & 1  & 15 &  1  \\
        \hline
         Vaccine 1 Effectiveness & 0.8 & 0.8  & 0.8 & 0.8  \\
        \hline 
         Vaccine 1 Availability & 450 & 450   & 100 & 100  \\
        \hline 
         Vaccine 2 Effectiveness & 0.6 & 0.6  & 0.6  & 0.6  \\    
        \hline 
         Vaccine 2 Availability & 450 & 450  & 700 &  700  \\  
        \hline 
       
        \multicolumn{5}{l}{\textbf{Scenario (Sc) 1: Health + Economy;
        Sc 2: Health; Sc 3: Economy}}
        \end{tabular}
        \label{experiments}
        \end{center}
    \end{table}

We had two types of rewards: 
\begin{itemize}
\item{Health of the individuals ($HRew$), given by ``Equation (\ref{eq1})". We did not use the deceased here since the numbers were very low in our model and we did not include diseased severe since they were all hospitalized in our model.}
\item{Economy reward function ($ERew$) was the negative of the sum of people below the poverty line ``Equation (\ref{eq2})".}
\end{itemize}
\begin{equation}
HRew = -1*(InfectedMild + Hospitalized)
\label{eq1}
\end{equation}

\begin{equation}
ERew = -1*(Below\_Poverty\_Line)
\label{eq2}
\end{equation}

\begin{equation}
TRew = HRew + \kappa*ERew
\label{eq3}
\end{equation}
The total rewards ($TRew$) is a combination of the health and the economic reward functions ``Equation (\ref{eq3})". The contribution of health and economic rewards is governed by a mixing factor $\kappa$. In each of the three scenarios, we vary the value of $\kappa$.
\begin{enumerate}
	\item{Scenario 1: Health \& Economy Priority} - Both the health and the economic rewards contributed equally to the total reward i.e. $\kappa = 1$.
	\item {Scenario 2: Health Priority} -The health rewards contributed more to the total reward i.e. $\kappa = 0.2$.
	\item {Scenario 3: Economy Priority} -The economy rewards contributed more to the total reward i.e. $\kappa = 5$.
\end{enumerate}

To maintain the integrity of the start and end dates, in all the scenarios, we optimized for start date and duration in DDPG and the endDate was computed as the sum of startDate and duration.

\begin{figure*}
\centering
\begin{subfigure}[t]{0.3\textwidth}
\centering
\includegraphics[width=\textwidth]{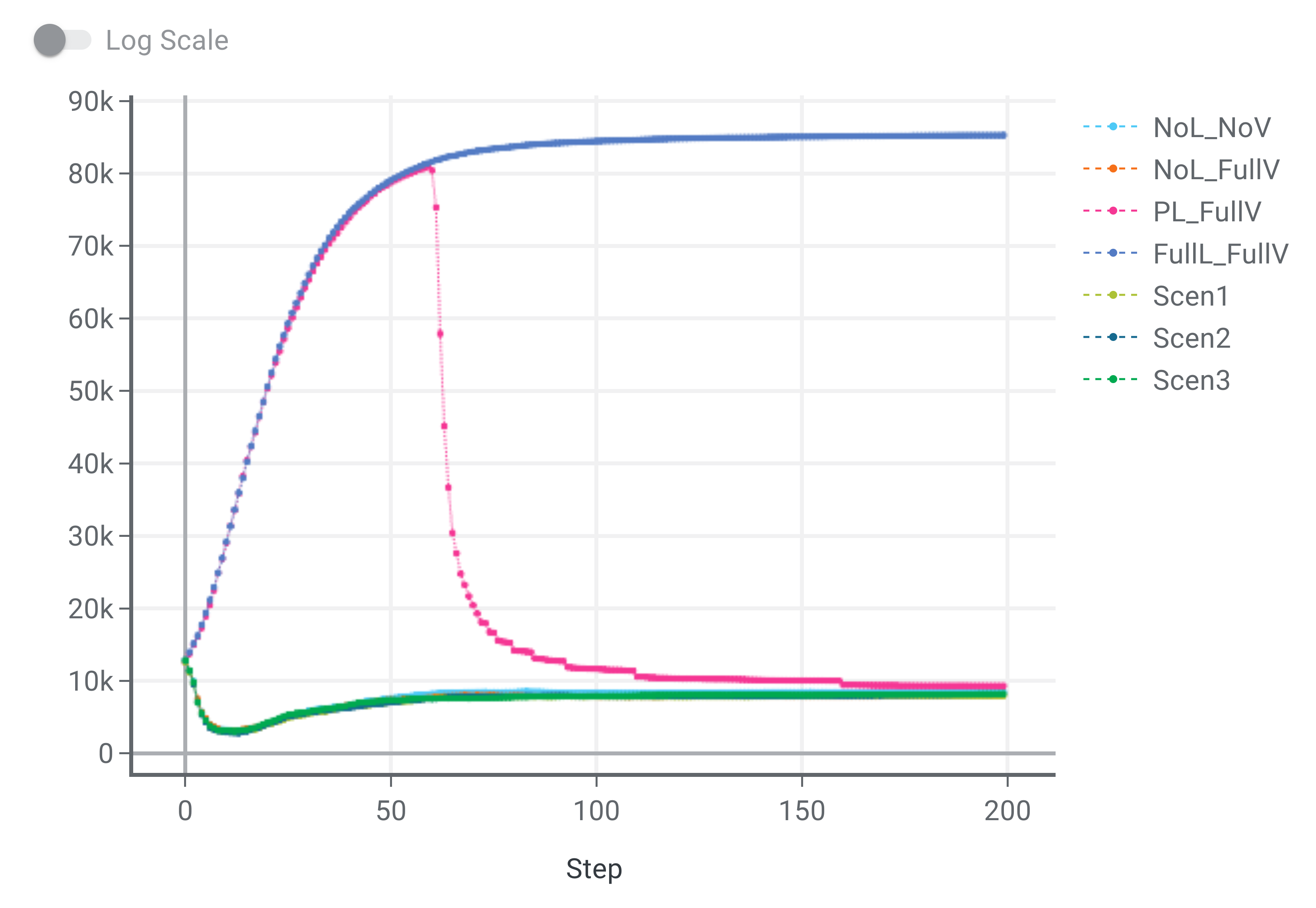} 
\caption{Below-Poverty-Line (yaxis) , x(axis number of days)} \label{1a}
\end{subfigure}
\hfill
\begin{subfigure}[t]{0.3\textwidth}
\centering
\includegraphics[width=\textwidth]{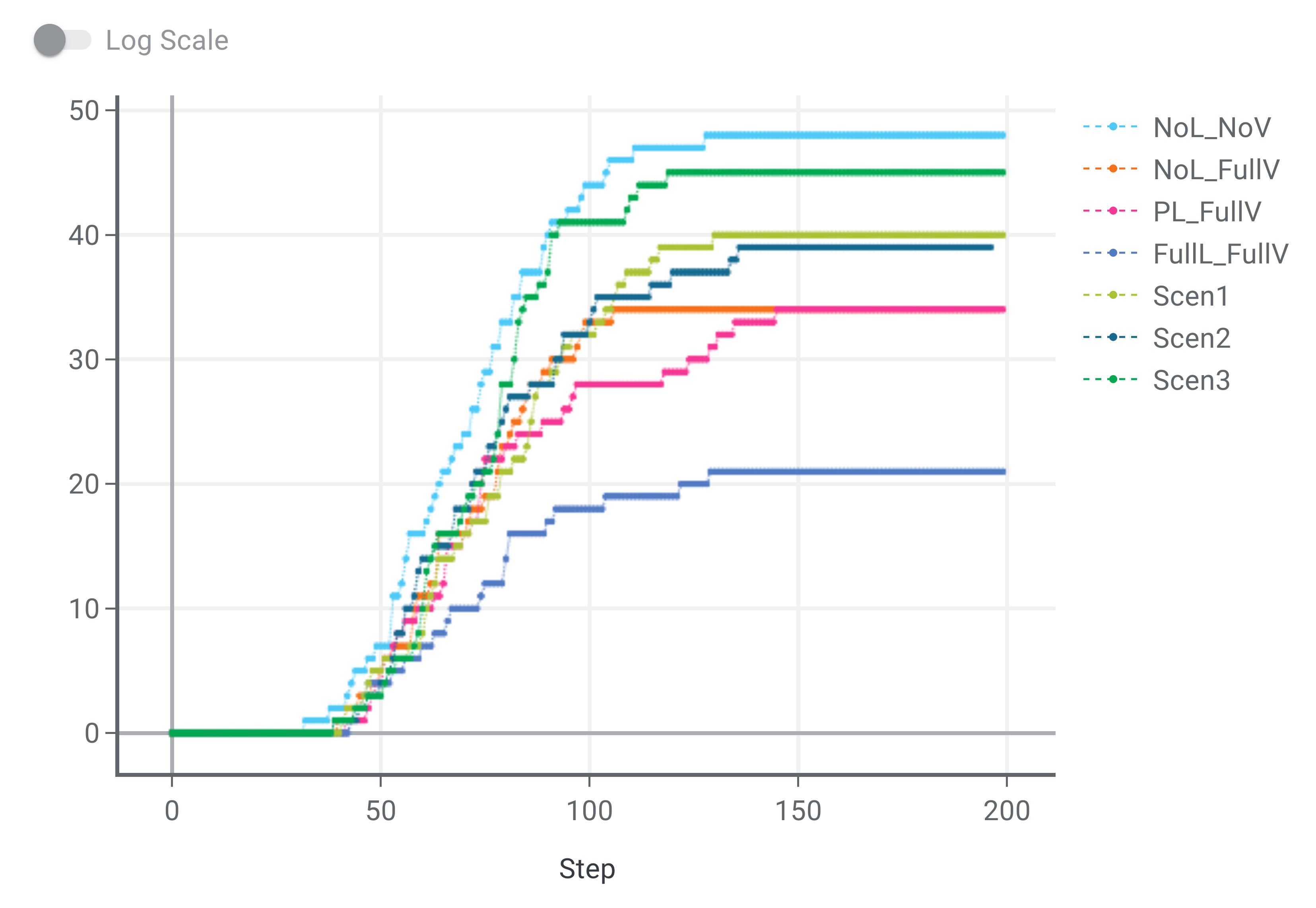} 
\caption{Deceased (yaxis) , x(axis number of days)} \label{1b}
\end{subfigure}
\hfill
\begin{subfigure}[t]{0.3\textwidth}
\centering
\includegraphics[width=\textwidth]{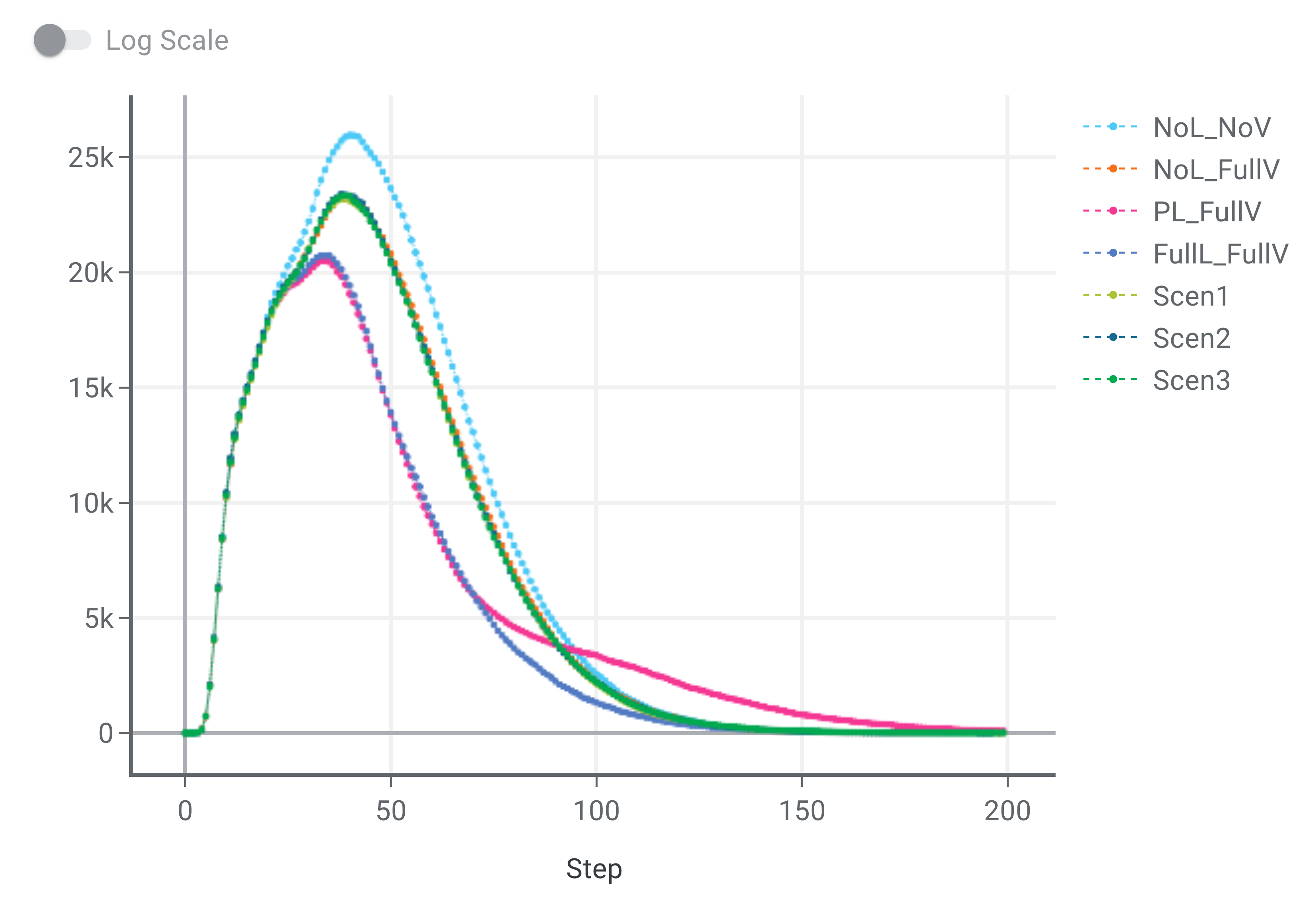} 
\caption{Total Infected (yaxis) , x(axis number of days)} \label{1c}
 \end{subfigure}

 \caption{Results of Experiment1: NoL refers to no lockdown and FullV refers to full vaccination through days 1-100. \subref{1a} all 3 scenarios have an economy score close to best-case outcome; \subref{1b} \& \subref{1c} deceased are between best and worst outcomes with the Scen2(Health) outperforming other 2 scenarios.}
\end{figure*}

\begin{figure*}
\centering
\begin{subfigure}[t]{0.3\textwidth}
\centering
\includegraphics[width=\textwidth]{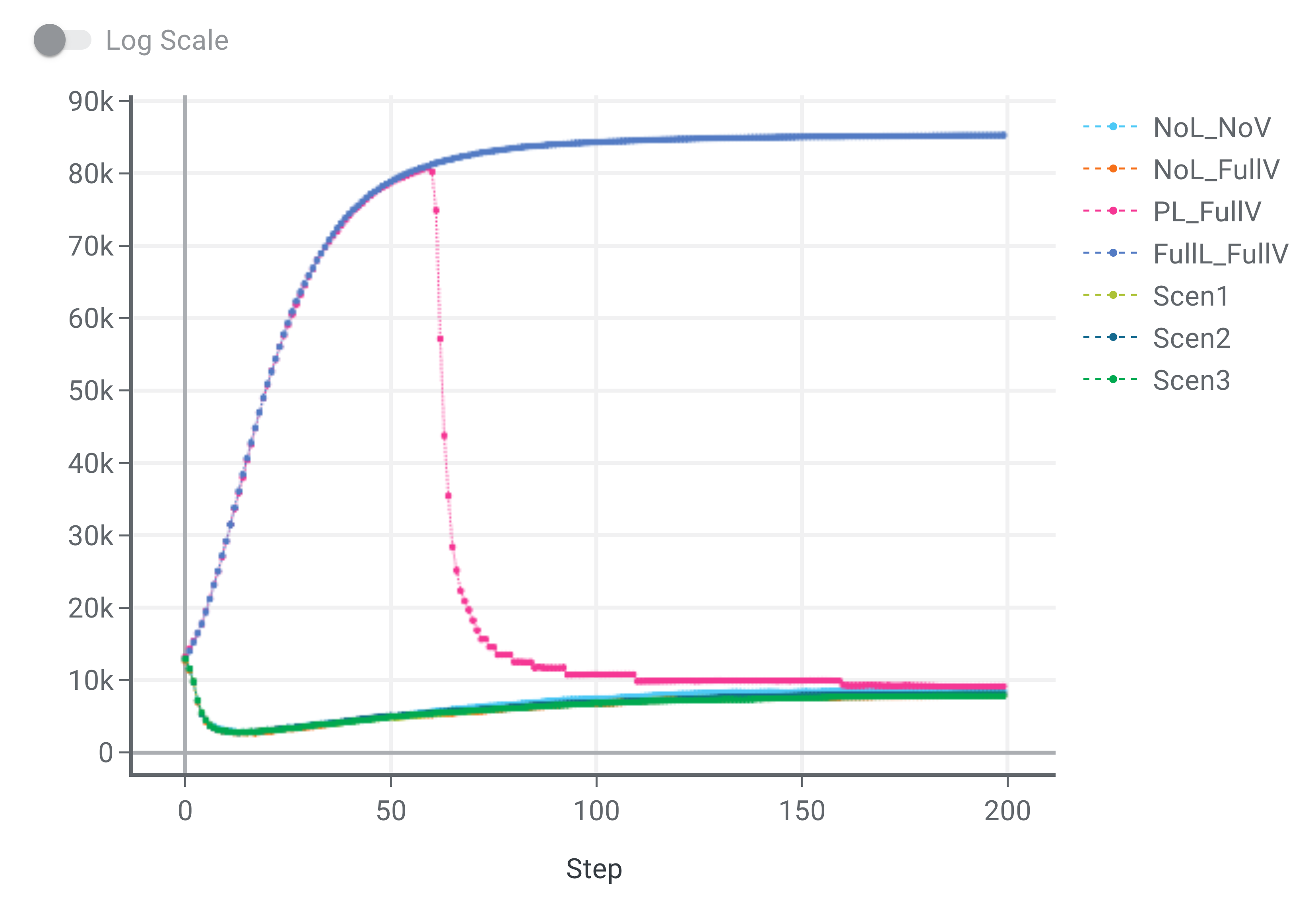} 
\caption{Below-Poverty-Line (yaxis) , x(axis number of days)} \label{2a}
\end{subfigure}
\hfill
\begin{subfigure}[t]{0.3\textwidth}
\centering
\includegraphics[width=\textwidth]{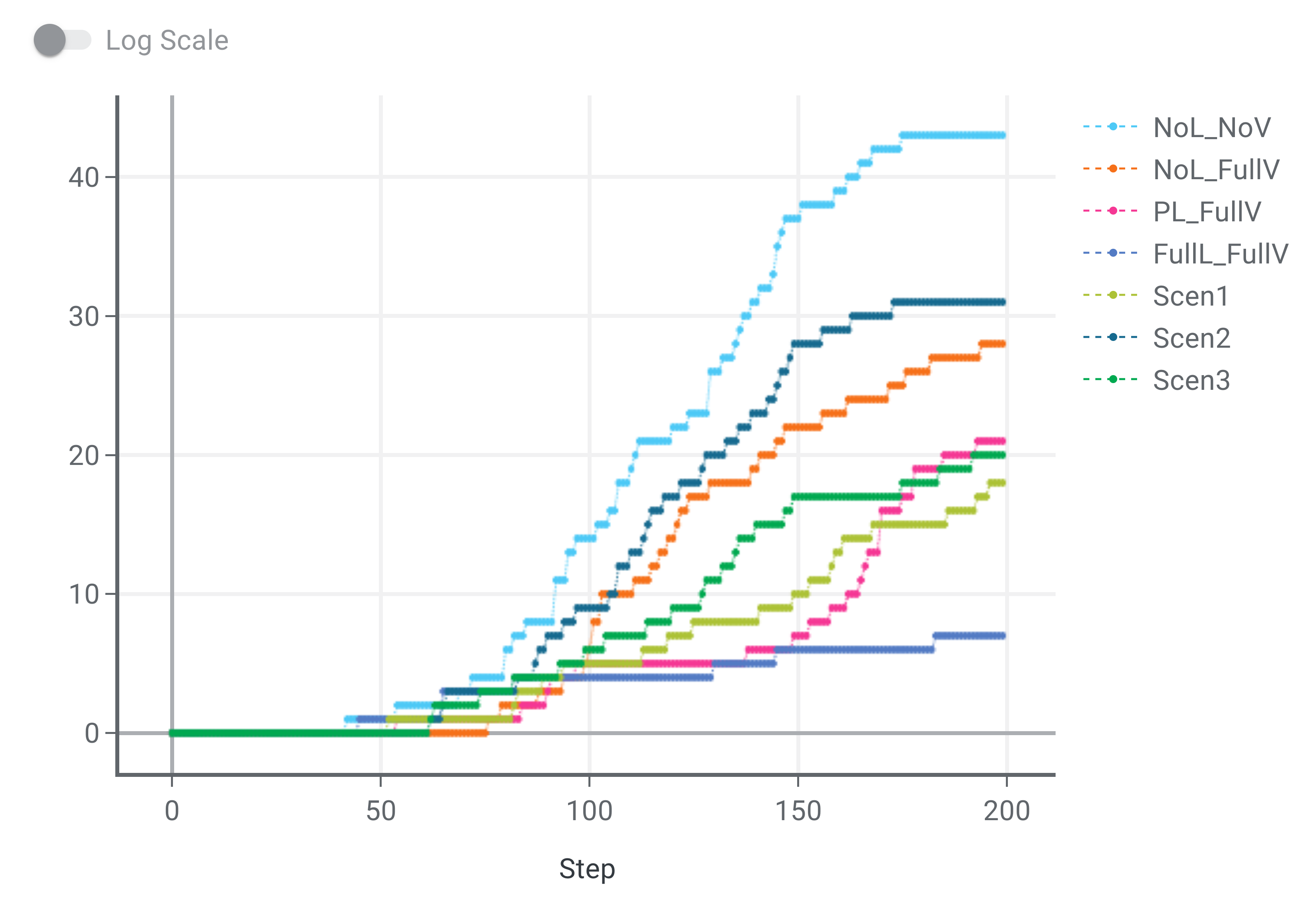} 
\caption{Deceased (yaxis) , x(axis number of days)} \label{2b}
\end{subfigure}
\hfill
\begin{subfigure}[t]{0.3\textwidth}
\centering
\includegraphics[width=\textwidth]{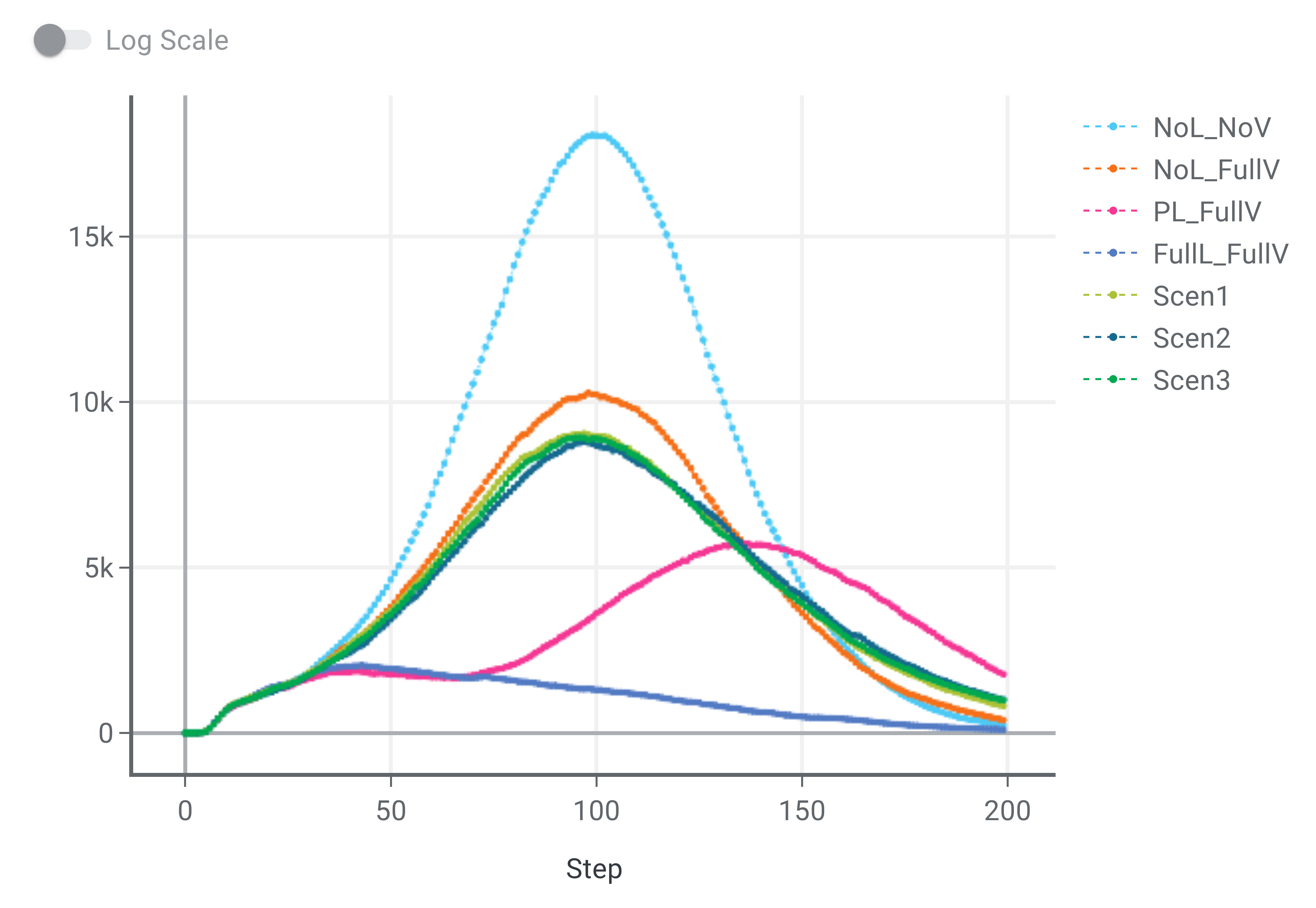} 
\caption{Total Infected (yaxis) , x(axis number of days)} \label{2c}
 \end{subfigure}

 \caption{Results of Experiment 2: NoL refers to no lockdown and FullV refers to full vaccination through day 1-100. \subref{1a} all 3 scenarios have economy score close to best-case outcome; \subref{1b} and \subref{1c} all 3 scenarios are between best and worst outcomes.}
\end{figure*}

\begin{figure*}
\centering
\begin{subfigure}[t]{0.3\textwidth}
\centering
\includegraphics[width=\textwidth]{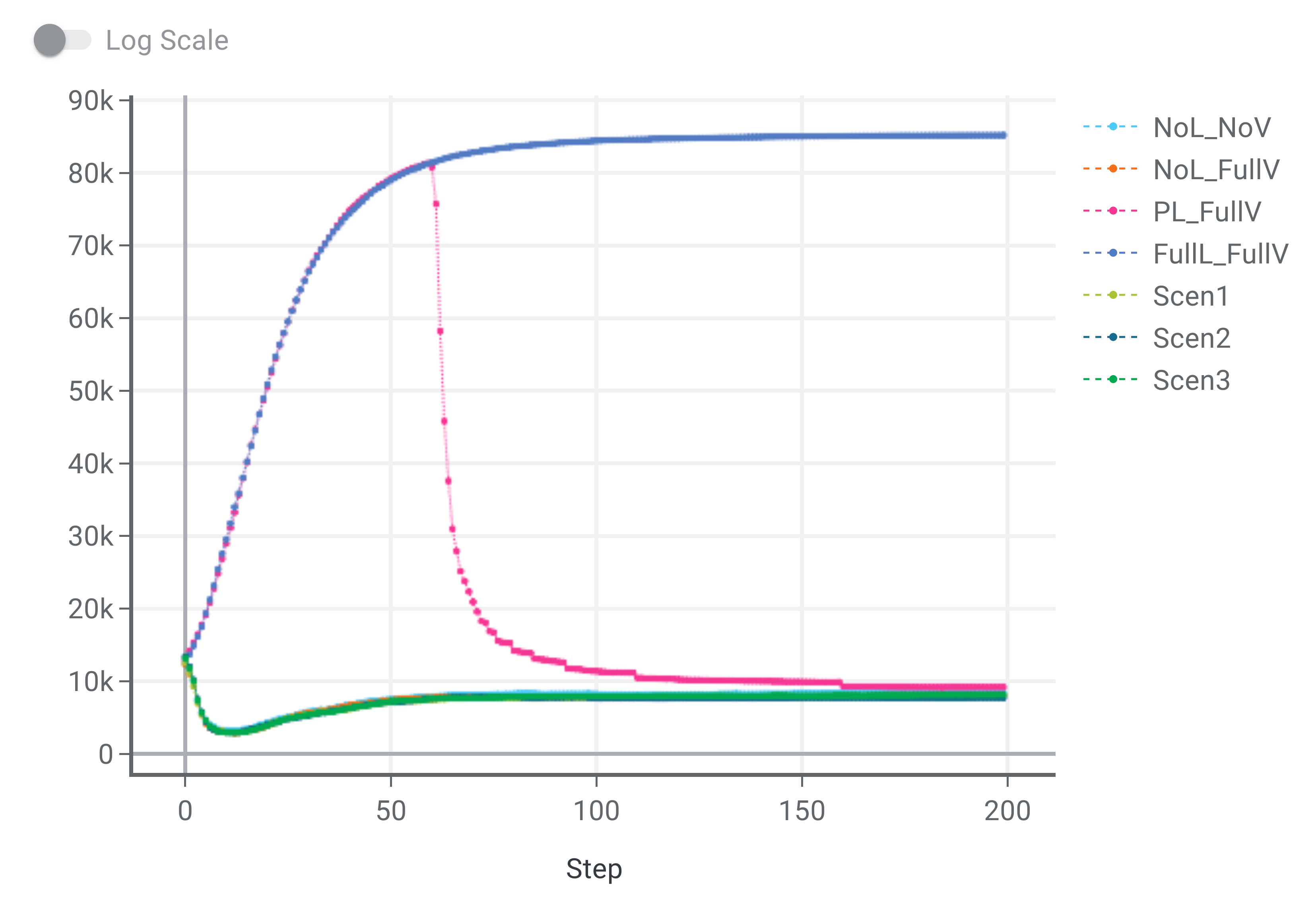} 
\caption{Below-Poverty-Line (yaxis) , x(axis number of days)} \label{3a}
\end{subfigure}
\hfill
\begin{subfigure}[t]{0.3\textwidth}
\centering
\includegraphics[width=\textwidth]{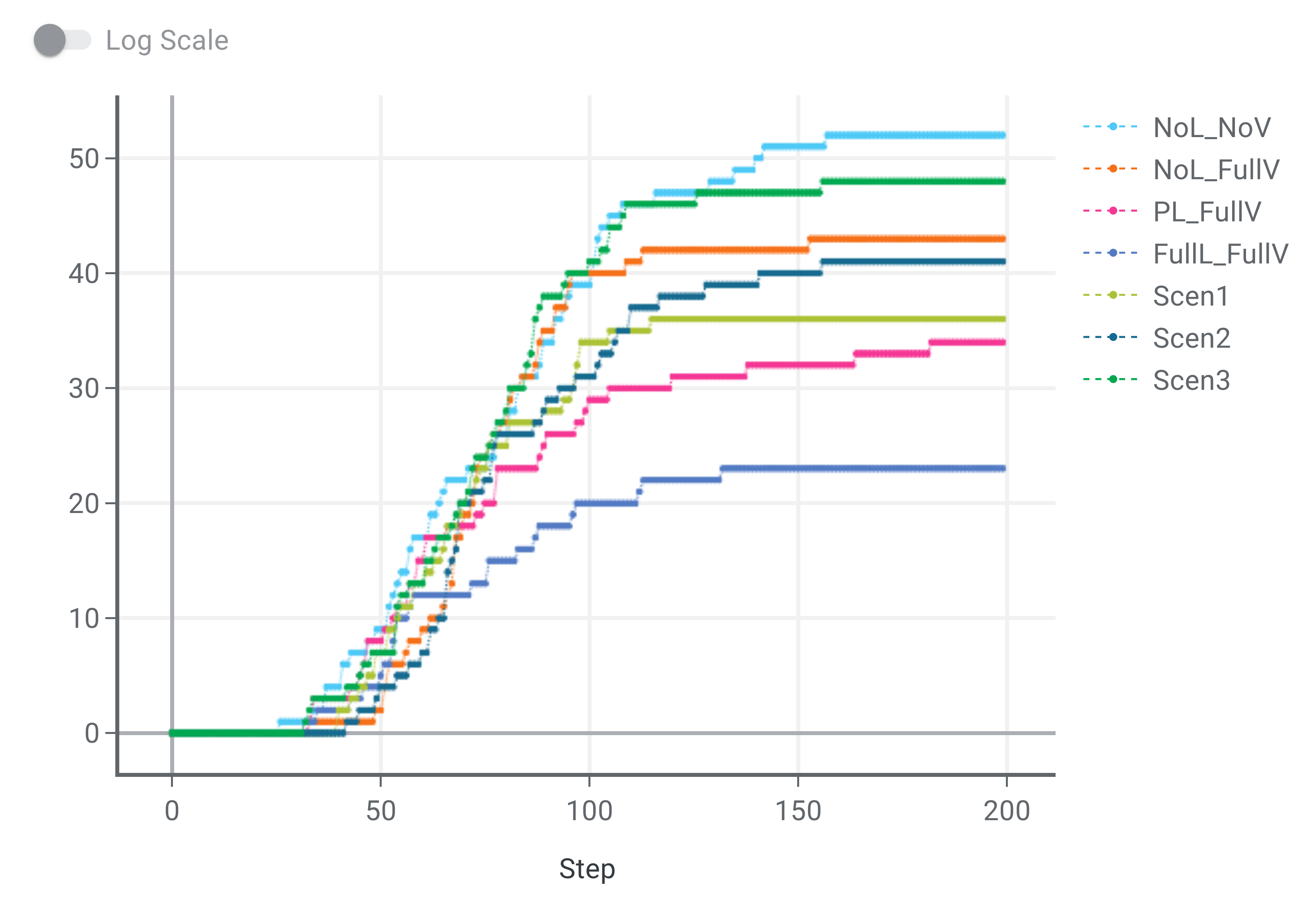} 
\caption{Deceased (yaxis) , x(axis number of days)} \label{3b}
\end{subfigure}
\hfill
\begin{subfigure}[t]{0.3\textwidth}
\centering
\includegraphics[width=\textwidth]{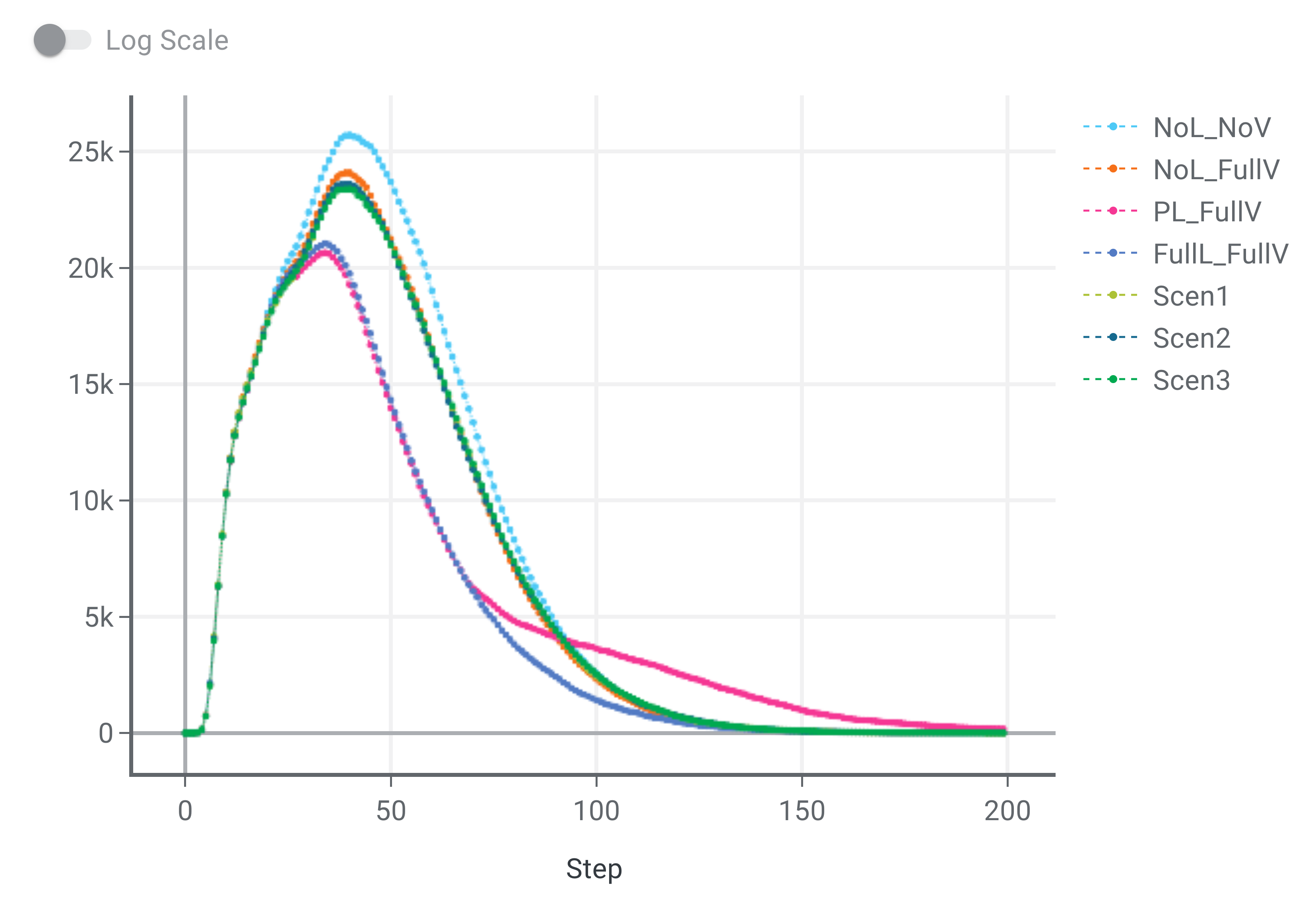} 
\caption{Total Infected (yaxis) , x(axis number of days)} \label{3c}
 \end{subfigure}

 \caption{Results of Experiment 3: NoL refers to no lockdown and FullV refers to full vaccination starting through 1-100. \subref{1a} all 3 scenarios have economy score close to best-case; \subref{1b} and \subref{1c} all 3 scenarios are between best and worst-case outcomes..}
\end{figure*}    

\begin{figure*}
\centering
\begin{subfigure}[t]{0.3\textwidth}
\centering
\includegraphics[width=\textwidth]{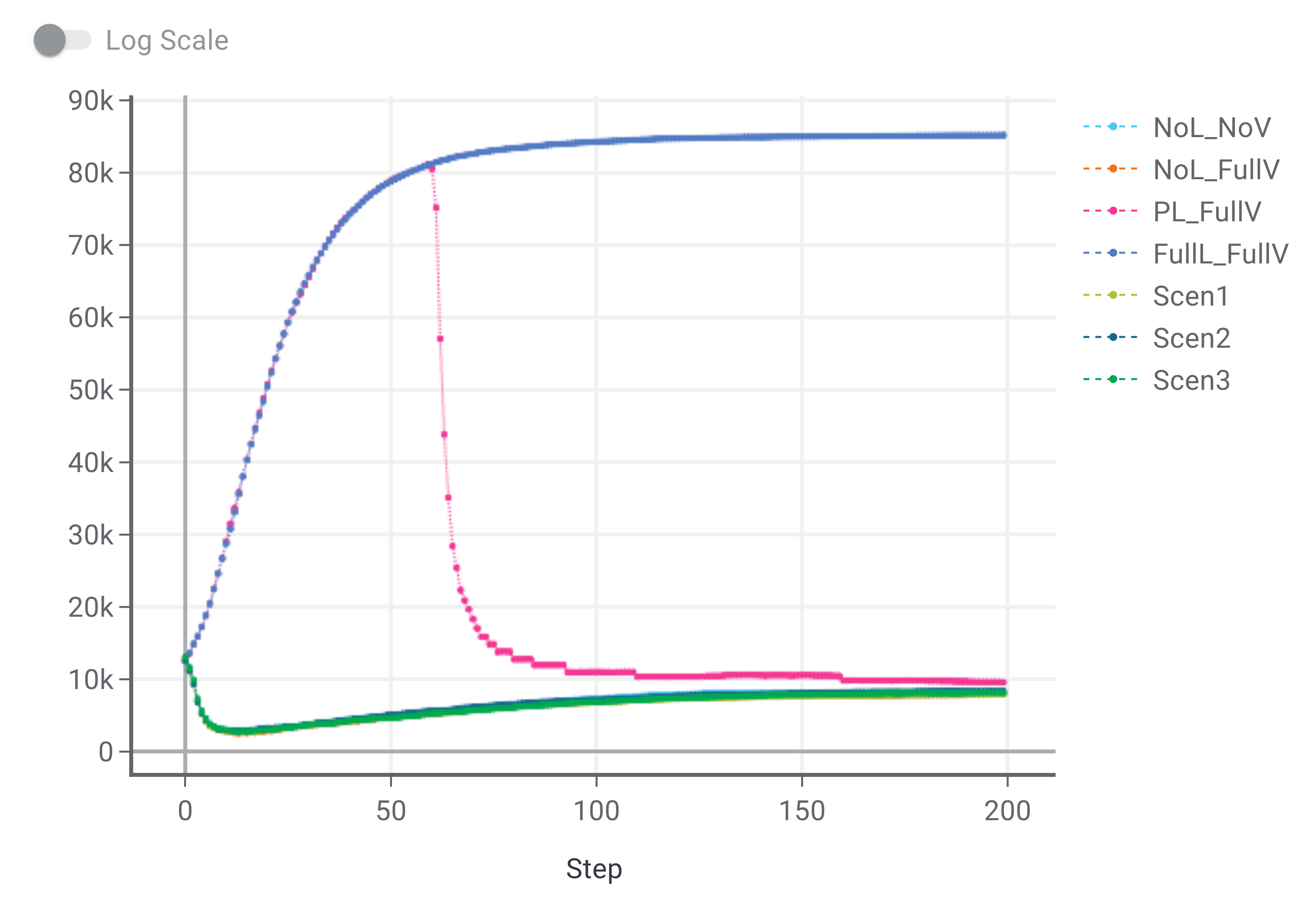} 
\caption{Below-Poverty-Line (yaxis) , x(axis number of days)} \label{4a}
\end{subfigure}
\hfill
\begin{subfigure}[t]{0.3\textwidth}
\centering
\includegraphics[width=\textwidth]{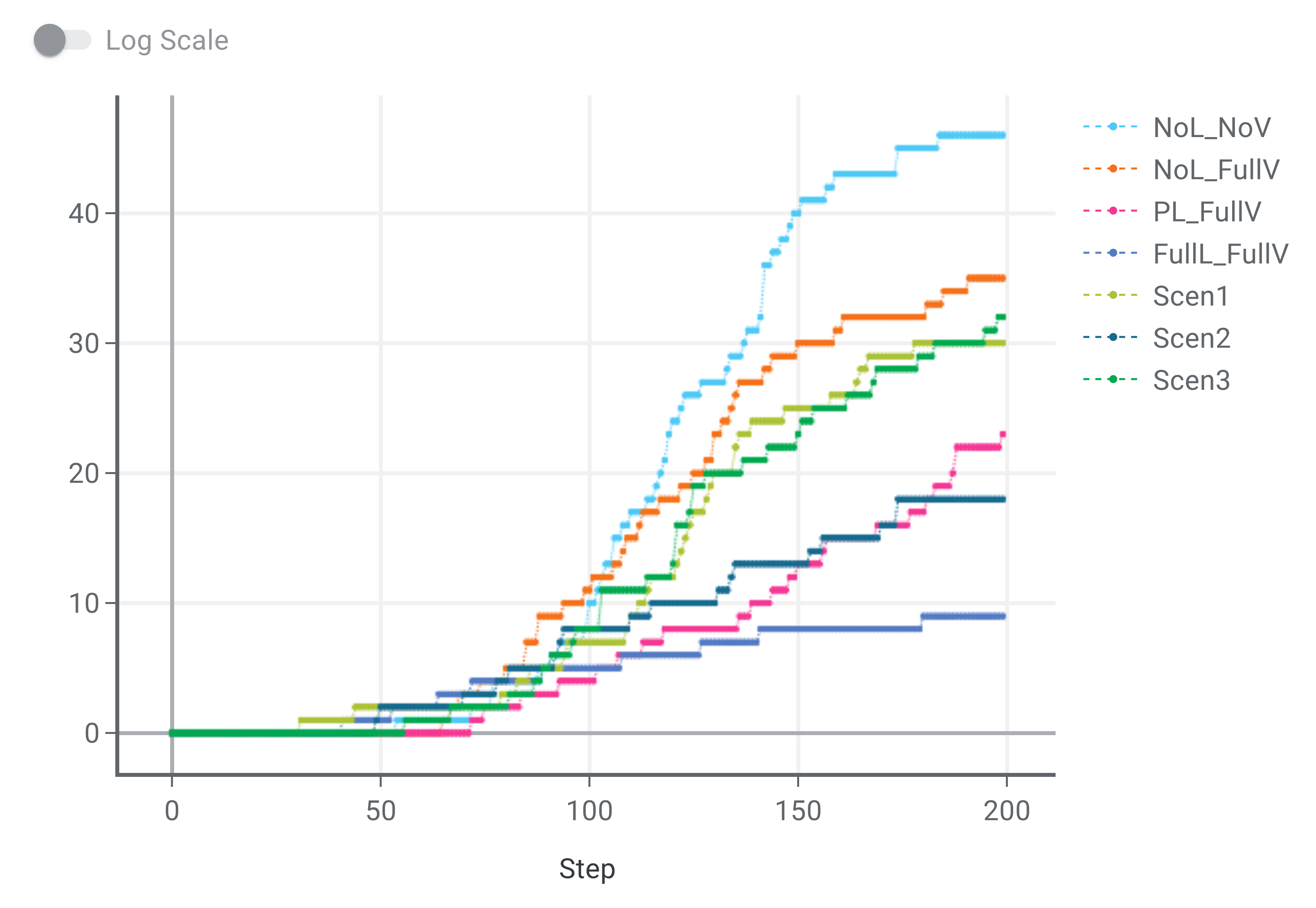} 
\caption{Deceased (yaxis) , x(axis number of days)} \label{4b}
\end{subfigure}
\hfill
\begin{subfigure}[t]{0.3\textwidth}
\centering
\includegraphics[width=\textwidth]{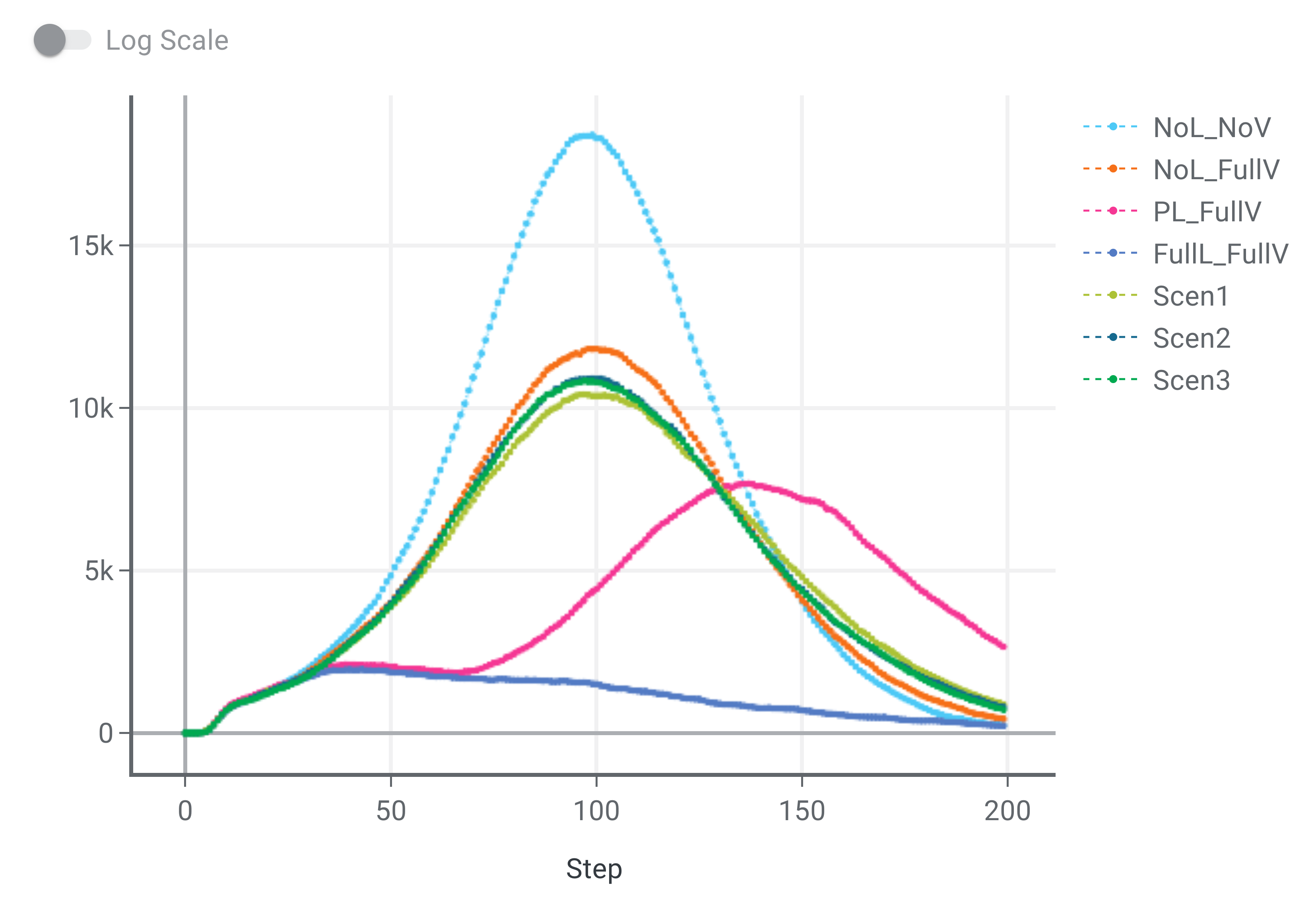} 
\caption{Total Infected (yaxis) , x(axis number of days)} \label{4c}
 \end{subfigure}

 \caption{Results of Experiment 4: NoL refers to no lockdown and FullV refers to full vaccination starting through 1-100. \subref{1a} all 3 scenarios to have an economy score close to best-case outcome; \subref{1b} and \subref{1c} all 3 scenarios are between best and worst-case outcomes.}
\end{figure*}

\subsection{DDPG Hyper-parameters}
As explained above we used an agent-based model (ABM) in our experiments. To account for the stochastic nature of the ABM model, we ran the experiments multiple times ($n = 2$) and took the mean of the experimental runs. For the DDPG, we used 150 training iterations, 10 burning steps, and 5 test repeats after every 10 steps. We tested the models for convergence before running the test setup. To compute the reward we used the sum of the max and the mean value of the states (i.e. the mild infection, hospitalization,  and the below-poverty-line count at each time step). The rationale was to flatten the curve and also to have few impacts on health and the economy. Based on our experiments, a combination of the mean and max gave the best results as opposed to the mean (which leads to short sharp peaks) or max (which leads to a prolonged infection) alone. 

The values for the other hyper-parameters are as follows:  
"seed": 0, "discount": 0.99, "tau": 5e-3, "expl\_noise": 0.1, "batch\_size": 32,

The models were run on a combination of CPUs and GPU clusters <specs>. Each simulation run took 2-3 min depending on the machine's specifications. Since we ran 2 repeats and 150 training iterations, each training cycle took between 5-8 hrs (accounting for some communication overheads). We did not have previous data to train on so we performed online training with the simulation runs.

\subsection{Results of Experiment 1: High Initial Infection \& Equal High-Efficiency, Low-Efficiency Vaccines}
Across all three scenarios (Figure \ref{1a}), the number of below-poverty-line individuals was very close to the best-case outcomes for the economy where no lockdown was imposed. Similarly, the total number of deaths, hospitalization, and total infections (Figures  \ref{1b}, \ref{1c}) though higher than the best-case outcome of ``full lockdown, day 1-100 vaccinations" were comparable with that of ``partial lockdown, day 1-100 vaccinations" and ``no lockdown, day 1-100 vaccination". All three reward scenarios behaved very similarly. Some possible explanations are that in our model, the infected individuals do not earn so prioritizing the economy also indirectly prioritizes health. As a result, unless we drastically deprioritize one over the other, the results are similar. The actions which were recommended by the agents in each scenario for all three scenarios were no lockdowns, no vaccination for ages 0-17, vaccinations days 0-41 for ages 18-59, and vaccinations days 0-100 for ages 60-99.

\subsection{Results of Experiment 2: Low Initial Infection \& Equal High-Efficiency, Low-Efficiency Vaccines}
Across all three scenarios (Figures \ref{2a}), the number of below-poverty-line individuals was very close to the best-case outcome for the economy, where no lockdown was imposed. Similarly, the total number of deaths (Figure \ref{2b}) though higher than the best-case outcomes of ``full lockdown, days 1-100 vaccinations" was comparable with that of ``partial lockdown, days 1-100 vaccination". Of the three reward  functions, Scenario 1(Health + Economy) and 3(Economy) showed the least amount of death. Across all scenarios, the total infections (Figure \ref{2c}) were significantly lower than the worst-case outcome and higher than the ``partial-lockdown, days 1-100 vaccination". The results of this experiment show that when an initial lockdown is imposed then that results in a shifting of the peak but has a significant impact on the economy. On the contrary vaccination from day 1 helps avert serious consequences to health and the economy. The actions which were recommended by the agents in each scenario for all three scenarios were no lockdowns, no vaccination for ages 0-17, vaccinations days 0-41 for ages 18-59, and vaccinations days 0-100 for ages 60-99.

\subsection{Results of Experiment 3: High Initial Infection \& Low volume High-Efficiency, High volume Low-Efficiency Vaccines}
Results of the experiment indicate that across all three scenarios, the number of below-poverty-line individuals (Figure \ref{3a}) was very close to the best-case outcome for the economy, where no lockdown was imposed. However, across all scenarios, the deaths, hospitalizations and total infections (Figures \ref{3b}, \ref{3c}) were lower than the worst-case outcome, higher than ``partial lockdown, days 1-100 vaccinations" and comparable to ``no lockdown, days 1-100 vaccinations". When the epidemic has a high start infection (15\%) the vaccinations do not eliminate the peak but however, can reduce the height of the peak. Our optimization performed comparably with full vaccination despite only performing limited vaccination. The actions which were recommended by the agents in each scenario for all three scenarios were no lockdowns, no vaccination for ages 0-17, vaccinations days 0-41 for ages 18-59, and vaccinations days 0-100 for ages 60-99.

\subsection{Results of Experiment 4: Low Initial Infection \& Low volume High-Efficiency, High volume Low-Efficiency Vaccines}
 The results for this experiment (Figures \ref{4a}, \ref{4b}, \ref{4c}) are similar to Experiment 2, but they also demonstrate that despite the availability of low-effectiveness vaccines, the infections, hospitalizations and deaths are only marginally higher if widespread vaccination is followed. The actions which were recommended by the agents in each scenario for all three scenarios were no lockdowns, no vaccination for ages 0-17, vaccinations days 0-41 for ages 18-59, and vaccinations days 0-100 for ages 60-99.

%% file: sections/discussion.tex
Across all the scenarios, no lockdowns were imposed, and similarly the age groups 60-99 were vaccinated for the entire duration of the simulation, the age groups 18-60 for nearly half of the simulation time, and the age 0-18 were almost never vaccinated. This is in line with the model where the elderly are highly susceptible and the younger populations are relatively immune. Despite not optimizing for death and total infection, the optimization could effectively control both these factors. Across the experiments we could also see that though the vaccination alone could not make a significant impact when the initial infection load was high, they were very effective when started earlier in the pandemic. The actions and conclusions drawn from our results are in-line with other studies on Covid Epidemics which promote high vaccinations for the elderly and then followed by the middle age group ~\shortcite{matrajt2021vaccine}. There however one point where our model differs from previous literature was that it advocated for simultaneous vaccinations of both age groups.  In addition, imposing lockdowns also had a severe impact on the economy and sometimes these were semi-permanent at best. These results indicate that optimization algorithms such as DDPG are very promising for use in policy informatics.

In terms of the rewards, the effects of economics played a higher role in the optimization than health. A possible explanation is that the impact of the interventions such as the lockdown was higher on the poverty line as opposed to the infections. Similarly, when people were infected, they did not earn, which again impacted the poverty level as a result even when the economy was favored, it indirectly helped health in our model. However, since the death was so less, their effect in the model may have been due to stochastic behavior. 

%% file: sections/conclusion.tex
The optimization of policy for epidemics and pandemics remains challenging especially in a multi-objective setting when there are conflicting priorities. Models for optimization have been largely limited in the scale of the population, model complexity, and the number of choices for interventions (discrete vs continuous). In this work, we illustrate the optimization of public health and economy using a DDPG-based reinforcement learning framework. We showcase four experimental settings on a minimalist model with different starting infections, vaccination efficiencies, and vaccination availability. The optimization framework gave optimal economy scores and health scores across all experiments and scenarios. We can conclude that the use of reinforcement learning frameworks such as DDPG shows promise in the field of policy informatics and such tools can effectively aid policymakers and computational epidemiologists.

Since this is a work-in-progress, we are working towards addressing the following limitations. We demonstrated our results using a single model with a few simple assumptions and a fixed agent behavior (using equations and some stochasticity). In the near future, we will expand on the model to include rational agents, with complex human and economic behavior. In addition, though we account for the stochastic nature of ABMs through multiple runs, we do not account for tolerance in the predicted actions when put into practice. We also do not employ explainability or techniques to illustrate trust in this study. In the near future, we also aim to extend our work to account for robustness, trust, and explainability. In addition, we will also extend our study to compare different RL algorithms with appropriate metrics.